%% file: root.tex
\documentclass[letterpaper, 10 pt, conference]{ieeeconf}  % Comment this line out if you need a4paper

\IEEEoverridecommandlockouts                              % This command is only needed if
\makeatletter
\let\NAT@parse\undefined
\makeatother
\usepackage[numbers, sort&compress]{natbib}
\usepackage{times}

\usepackage{tikz}
\usepackage{graphicx} % for pdf, bitmapped graphics files
\usepackage{amsmath} % assumes amsmath package installed
\usepackage{dblfloatfix}
\usepackage{makecell}

\usepackage{amssymb}  % assumes amsmath package installed
\usepackage{amsfonts}
\usepackage{subfigure}
\usepackage{ctable}
\usepackage{makecell}

\usepackage[titlenumbered,linesnumbered,ruled,vlined]{algorithm2e}

\usepackage{fancyvrb}
\usepackage{xcolor}
\usepackage{verbatimbox}

\usepackage{caption}
\usepackage[numbers]{natbib}
\usepackage{multicol}
\usepackage[bookmarks=true]{hyperref}

\usepackage{amsthm}
\theoremstyle{definition}
\usepackage{src/irap_acronyms}
\usepackage{src/irap_math}
\usepackage{src/irap_SIunits}
\usepackage{src/irap_misc}

\usepackage{soul,color}
\usepackage{verbatim} % for multiline comment
\usepackage{multirow}
\usepackage{cleveref}

\usepackage{lipsum}

\usepackage{textcomp}
\usepackage{lipsum,multicol}
\usepackage{xcolor}

\DeclareMathOperator*{\argmin}{argmin}

\definecolor{myblue}{RGB}{190,255,255}
\sethlcolor{myblue}

\let\oldnl\nl% Store \nl in \oldnl
\newcommand{\nonl}{\renewcommand{\nl}{\let\nl\oldnl}}% Remove line number for one line

\begin{document}
	
	\title{\LARGE \bf
		ViViD++ : Vision for Visibility Dataset
	}
	
	\author{Alex Junho Lee$^{1}$, Younggun Cho$^{2}$, Young-sik Shin$^{3}$, Ayoung Kim$^{4}$ and Hyun Myung${}^{5*}$
		%\thanks{*\hl{Need ack.}}% <-this % stops a space
		\thanks{$^{1}$Alex Junho Lee is with the Department of Civil and Environmental Engineering, \acs{KAIST}, Daejeon, S. Korea.
			{\tt\small alex\_jhlee@kaist.ac.kr}}%
		\thanks{$^{2}$Y. Cho is with the Department of Electrical Engineering, Inha University, Incheon, S. Korea. {\tt\small yg.cho@inha.ac.kr}}%
		\thanks{$^{3}$Y. Shin is with Korea Institute of Machinery and Materials, Daejeon, S. Korea.
			{\tt\small yshin86@kimm.re.kr}}%
		\thanks{$^{4}$A. Kim is with the Department of Mechanical Engineering, SNU, Seoul, S. Korea {\tt\small ayoungk@snu.ac.kr}}
		\thanks{$^{5}$H. Myung is with the School of Electrical Engineering, \acs{KAIST}, Daejeon, S. Korea.
			{\tt\small hmyung@kaist.ac.kr}}%
		\thanks{This work was supported in part by Institute of Information \& Communications Technology Planning \& Evaluation (IITP) grant funded by Korea government (MSIT) (No.2020-0-00440, Development of Artificial Intelligence Technology that Continuously Improves Itself as the Situation Changes in the Real World), and in part by ``Indoor Robot Spatial AI Technology Development" project funded by KT (KT award B210000715).
		The students are supported by the BK21 FOUR from the Ministry of Education (Republic of Korea).}
	}

	\maketitle
	\thispagestyle{empty}
	\pagestyle{empty}
	
	%%%%%%%%%%%%%%%%%%%%%%%%%%%%%%%%%%%%%%%%%%%%%%%%%%%%%%%%%%%%%%%%%%%%%%%%%%%%%%%%
	
	\input{src/abstract.tex}
	\input{src/introduction.tex}
	\input{src/related_works.tex}
	\input{src/environment.tex}
	\input{src/dataset.tex}

	\input{src/conclusion.tex}
	
	%% Use plainnat to work nicely with natbib.
	%\bibliographystyle{plainnat}
	% % \bibliographystyle{IEEEtranN} % not IEEEtran, but IEEEtranN for using citeauthor
	% \bibliography{references}
	\renewcommand*{\bibfont}{\small}
	\bibliographystyle{IEEEtranN} % not IEEEtran, but IEEEtranN for using citeauthor
	\bibliography{string-short,src/references}
	
	% for conf (IROS, ICRA)
	% \renewcommand*{\bibfont}{\small}
	% \bibliographystyle{IEEEtranN} % not IEEEtran, but IEEEtranN for using citeauthor
	% \bibliography{string-short,references}
	% \newcommand{\bioshot}[1]{\includegraphics[width=1in,height=1.25in,clip,keepaspectratio]{#1}}
	
	\vfill
	
\end{document}

%% file: src/abstract.tex
\begin{abstract}

In this paper, we present a dataset capturing diverse visual data formats that
target varying luminance conditions. While RGB cameras provide
nourishing and intuitive information, changes in lighting conditions potentially
result in catastrophic failure for robotic applications based on vision sensors.
Approaches overcoming illumination problems have included developing more robust
algorithms or other types of visual sensors, such as thermal and event cameras.
Despite the alternative sensors' potential, there still are few datasets with
alternative vision sensors. Thus, we provided a dataset recorded from alternative vision
sensors, by handheld or mounted on a car, repeatedly in the same space but in different
conditions. We aim to acquire visible information from co-aligned
alternative vision sensors. Our sensor system collects data more independently from
visible light intensity by measuring the amount of infrared dissipation, depth by
structured reflection, and instantaneous temporal changes in luminance. We provide
these measurements along with inertial sensors and ground-truth for developing
robust visual SLAM under poor illumination. The full dataset is available
at: \url{https://visibilitydataset.github.io/}

\end{abstract}

%% file: src/introduction.tex
\section{Introduction}

With recent interests in autonomous navigation, a robot's ability to achieve
localization and recognize the surroundings has been a critical feature for
mobile applications. To solve the autonomous navigation problem in the real world,
cameras have been widely used for their cost-effectiveness and intuitiveness.
Numerous studies based on images have focused on developing robust visual
\ac{SLAM} algorithms to cope with real-world disturbances, such as lighting
and motion variances. The scene's visual deviation arises from the natural
and artificial illumination changes. To overcome the visual disturbances and
develop robust visual \ac{SLAM}, numbers of datasets covering variations were introduced.

\input{src/figtex/sensors.tex}

Public datasets provide large environmental variations from cameras.
Some of them provide measurements from indoor \cite{sturm12iros,Burri25012016,jeon2021run}
and the others arrange outdoor scenes~\cite{Geiger2012CVPR,Cordts2016Cityscapes,jjeong-2019-ijrr}
providing a large scale of sensor measurements. Synthetic datasets
\cite{handa:etal:ICRA2014,mccormac2017scenenet,li2018interiornet} were also provided for diversity.

However, achieving robust visual \ac{SLAM} under low visibility still remains
challenging. For typical cameras that collect information by integrating photons
during the exposure time, the captured image is concentrated on the lighting more than the object.
Thus, the ideas of collecting information from visual domains other than visible light
intensity have been introduced. The alternative vision sensors have advantages over typical cameras.
For instance, the thermal cameras could capture infrared radiation, and the event cameras \cite{berner2013240}
could detect temporal changes. These special abilities make the sensor measurement to be
more independent from external lighting and motion conditions. For datasets including
alternative vision sensors, event datasets
\cite{weikersdorfer2014event,mueggler2017event,zhu2018multivehicle,Delmerico19icra,fischer2020event,gehrig2021dsec}
and thermal datasets \cite{maddern2012towards,8293689} were publicly released.

In this paper, we present a dataset for developing robust visual \ac{SLAM} in the real world by providing:
\begin{itemize}
	\item the first dataset to enclose information from multiple types of aligned alternative vision sensors;
	\item multi-sensory measurements with ground-truth from external positioning system and generated from SLAM;
	\item wide range of environments in indoor and outdoor, recorded from multiple platforms.
\end{itemize}

%% file: src/figtex/sensors.tex
%FIGURE
\setcounter{figure}{0}
\begin{figure}[!t]
	\centering
	\captionsetup{font=footnotesize}
	\includegraphics[width=\columnwidth]{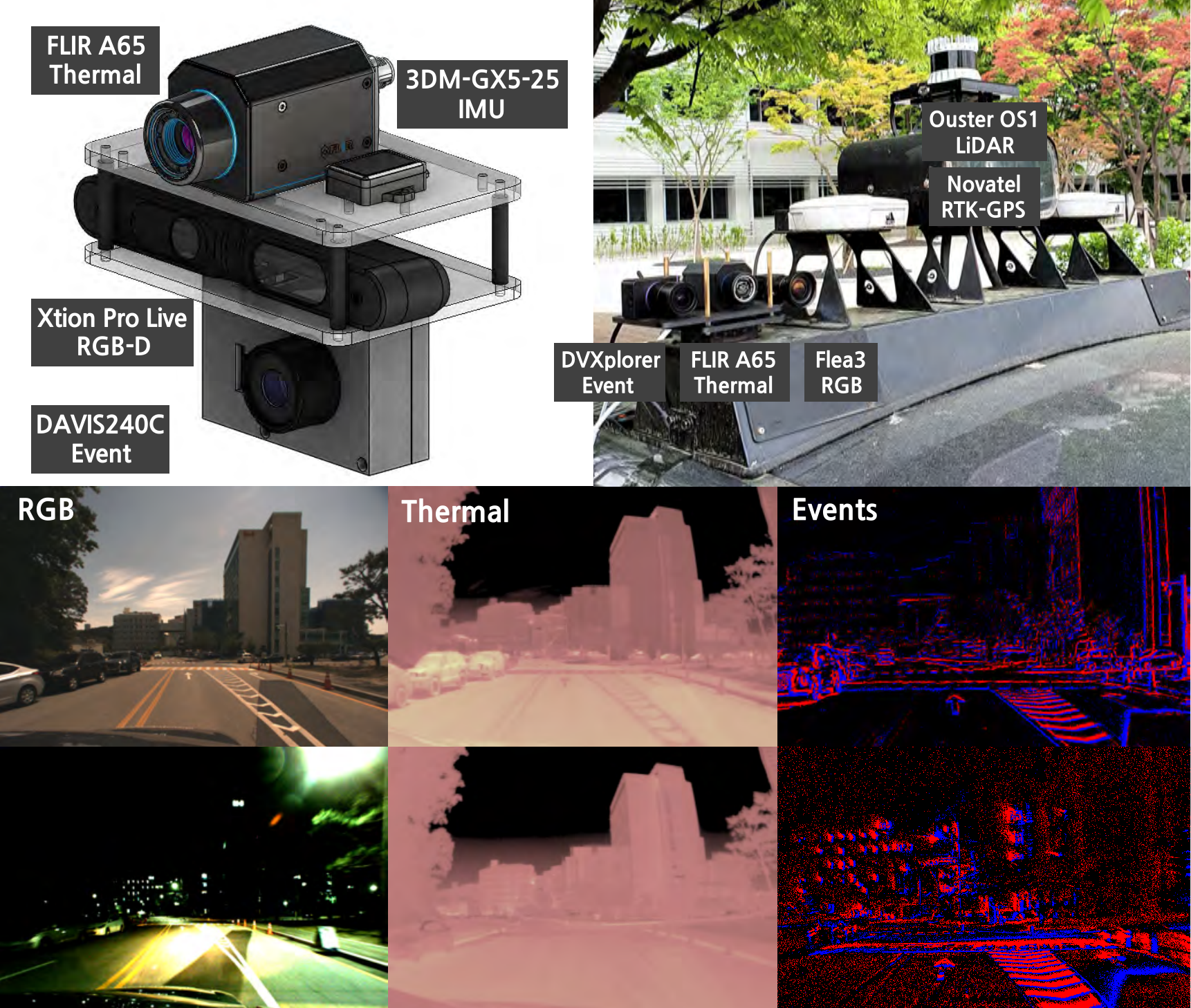}
	\label{fig:visibilisy}
	\vspace{-4mm}
	\caption{An overview of the sensor system and the dataset. We set up different sensor
		configurations for handheld (upper left) and driving (upper right).
		Sensor systems include RGB, thermal, events, depth, and interial measurements.
		Each sensor is indicated with the letter box. Each sensor's visibility differences
		are displayed in the lower row. Relative to RGB, thermal and
		event show less variance to the illumination condition.}
	\label{fig:sensors}
	\vspace{-4mm}
\end{figure}

%% file: src/related_works.tex
\section{Related Works}

\subsection{Related Datasets}
Vision sensors produce a wide range of information, but they are easily influenced by
external factors such as illumination and motion. As a result, a visual SLAM must be
evaluated in a wide range of scenarios. Existing datasets cover a wide range of scenarios
and include a large number of sensors, so as to develop robust SLAM algorithms. Various
datasets, encompassing a wide range of environments, were released, including a wide
range of visual and structural scenes. These datasets served as benchmark baselines and
enabled the advancements for research.

In TUM RGB-D \cite{sturm12iros}, image and depth data were collected, with ground-truths
from the motion capture system. To gradually test the robustness of visual sensors over movement,
sequences were separated by the motion speed. Similarly, in EuRoC \cite{Burri25012016},
images from stereo cameras were measured along with \ac{IMU} and ground-truth.
They separated the sequences with the difficulty of camera motion to test the robustness
over kinematic variances. Further in SceneNet~\cite{mccormac2017scenenet}, photo-realistic
videos, including motion blur and nonlinear camera response were provided with ground-truth.
In InteriorNet~\cite{li2018interiornet}, a large scale of synthetic environment with various
customizable parameters, i.e., texture, illumination, artificial lights, motion blur, etc.,
were presented with a wide choice of sensors such as depth, fisheye, and event cameras.
In NCLT \cite{ncarlevaris-2015a} and TUM monoVO \cite{engel2016photometrically},
a large scale of indoor and outdoor data with light and weather changes, was provided for long-term
visual \ac{SLAM}. Including all the variances mentioned above in urban environment,
datasets such as KITTI~\cite{Geiger2012CVPR}, Cityscapes~\cite{Cordts2016Cityscapes}, and
Complex urban dataset \cite{jjeong-2019-ijrr} were suggested.

Although many studies already covered a wide range of real-world noises, the
problem of classical cameras, which are highly dependent on external light,
remains. To overcome the appearance change and motion problem of conventional cameras,
datasets using alternative vision sensors, such as thermal and event cameras, were also introduced.

\subsection{Alternative Vision Datasets}

Unlike conventional cameras that capture intensity from the visible light
spectrum, thermal cameras capture infrared intensity from object surfaces.
The infrared measurements are generally more dependent on the object's temperature than
external light; thermal cameras can capture the scene information independently from
external light conditions. Further, the event cameras \cite{berner2013240}
are visual sensors that capture relative luminance changes over time, not the absolute
intensity. To code the instant light changes while maintaining low latency, the event
camera uses a particular type of data representation: streams of tuples, not the
sequence of images. Thus, the sensors' output characteristics are distinctive
from typical cameras. The event cameras require unique methods to process data,
and several datasets have been released.

\input{src/figtex/related_datasets.tex}

In early stages, \cite{weikersdorfer2014event} combined eDVS sensor with depth sensor and
released data in indoor sequences. Later the Event-Camera Dataset and Simulator
\cite{mueggler2017event} was released, enabling the simulated events and widening
the selection of test environments.

MVSEC \cite{zhu2018multivehicle} is a dataset with various lighting and motion
measured with stereo event camera and inertial sensors. The dataset contains a large
variety of condition changes, suggesting the utilization of the event camera for high-speed
motion estimation. However, the dataset does not include repeated measurements in
outdoor environments. In UZH-FPV \cite{Delmerico19icra}, aggressively moving drones
were used to provide event camera data for extreme motion cases. Also, a large-scale event
camera dataset was presented in \cite{gehrig2021dsec}, covering most of applicable
environments.

To investigate the appearance change problem with the event camera, in the Brisbane-Event-VPR
dataset \cite{fischer2020event}, the event camera was mounted on the car and acquired data
along the same trajectory at different times. The dataset provides a large appearance change
due to sun elevation changes. Especially in midnight sequences, street lights
largely affect the cameras, and VPR becomes nearly impossible, opening a problem to robustly
run event-based algorithms under both motion and appearance changes.

Also for thermal cameras, datasets covering illumination changes have been released.
In \cite{maddern2012towards}, the handheld rig of RGB, thermal, and \ac{GPS} receiver was
constructed, and collected data at different times. For object detection in thermal images,
Then, KAIST Multi-Spectral Day/Night Dataset \cite{8293689} have presented a sensor system including
stereo RGB, LiDAR, and thermal camera for SLAM purposes. Their data were collected with \ac{GPS} at
a large scale and along with daytime temperature changes.

As listed above, several datasets have provided environmental variances with different
types of sensors in various environments. However, as in \tabref{tab:related}, none of the datasets
include multi-sensor configurations for solving motion disturbances and illumination conditions
together, which are two significant variances in the real world. We insist that both sensors should
be utilized for complementing each other because thermal cameras are more robust by less dependency
on external light sources, and event cameras are better for extreme motion cases.
Therefore we compose a dataset with multiple types of visual information, at different
wavelengths and temporal intensity difference, along with depth measurements and ground-truth for
SLAM.

%% file: src/figtex/related_datasets.tex
\begin{table}[t]
	\centering
	\vspace{5mm}
	\captionsetup{font=footnotesize}
	\caption{Comparison with previous alternative vision datasets}
	\vspace{-2mm}
	\setlength{\tabcolsep}{4pt}
	\begin{tabular}{lccccc}
		\hline
		\multirow{3}{*}{Dataset} &&&Sensor&& \\ \cline{2-6}
                                             & Intensity  & Depth      & Thermal    & Event      & GT         \\  \cline{1-6}
  Mueggler et al.  \cite{mueggler2017event}  & \checkmark &            &            & \checkmark & \checkmark \\
  MVSEC            \cite{zhu2018multivehicle}& \checkmark & \checkmark &            & \checkmark & \checkmark \\
  UZH-FPV          \cite{Delmerico19icra}    & \checkmark &            &            & \checkmark & \checkmark \\
  DSEC             \cite{gehrig2021dsec}     & \checkmark & \checkmark &            & \checkmark & \checkmark \\
  Fischer et al.   \cite{fischer2020event}   & \checkmark &            &            & \checkmark & \checkmark \\
  Maddern et al.   \cite{maddern2012towards} & \checkmark &            & \checkmark &            & \checkmark \\
  Choi et al.      \cite{8293689}            & \checkmark & \checkmark & \checkmark &            & \checkmark \\\hline
ViViD++ (proposed)                           & \checkmark & \checkmark & \checkmark & \checkmark & \checkmark \\ \hline
	\end{tabular}
	\vspace{-8mm}
	\label{tab:related}
\end{table}

%% file: src/environment.tex
\input{src/format_table.tex}

\section{Environment}

\subsection{Sensors and Data types}

We setup different types of vision sensors: RGB, thermal, and event.
As mentioned in \figref{fig:sensors}, we configured different sensor
systems for handheld and driving sequences. In both scenarios, the three types of
visual sensors were included, but details differ. IMU was installed for handheld,
and the depth sensor was changed from depth camera to LiDAR for outdoor because
we observed that the depth measurements of a structured light depth camera were
unreliable on the outside. The dataset is provided with the binary format in rosbag.
The rostopic names, sensor model, and specifications are listed in \tabref{tab:types}.
The timestamp from the clock running on each sensor frame, was initialized to rostime
at startup and recorded in the header of each message.
For the thermal camera, FLIR A65 was used with the raw format of 16 bits instead of the typical 8 bits.
Note that the actual bits used for image demonstration are 14 bits due to two empty bits
in the front. The non-uniformity-correction (NUC) was set to manual and done at the beginning
of each sequence but not during the experiment. For the event camera, different
sensors were used for each setup.
DAVIS240C, which produces events and intensity images, was
used for handheld sequences. And DVXplorer, with higher spatial resolution but without
intensity images, was used for driving sequences. The LiDARs used for each system also differs.
Velodyne LiDAR with 16 channels and 100m range was used for handheld sequences,
and Ouster LiDAR with 64 channels and 120m range was used for driving sequences.

%--------------------------------------------------------------------------%
\subsection{Ground-Truth Generation}
\label{sec:groundtruth}

\subsubsection{Handheld indoor}
A cortex motion capture system KARPE (KAIST Arena with Real-time
Positioning Environment) \cite{karpe} was used to track the sensor system's
6 Degree of Freedom (DoF) pose. The system uses multiple infrared strobes to
track the reflected light from the markers, assigning marker IDs to each track.
The ID given to each marker did not change during the experiment,
and the minimum number of markers to compose a 6DoF tracking was three.
To ensure a robust track, we installed additional markers over the minimum number.
As the motion capture system only provides the position of each marker, not the rotation,
we defined the 6DoF marker frames from the tracked position of markers and provided
them as the ground-truth.

\subsubsection{Handheld outdoor}
We have generated poses for handheld outdoor sequences because the motion capture system
was only available with preinstalled indoor equipment, and the GPS receiver was not
able to solve the precise location in the experiment site. Thus as an evaluation
baseline, we used the pointcloud from scans of VLP-16 attached to the top of the sensor jig and
generated pose with LOAM variants~\mbox{\cite{legoloam2018}\cite{aloamgithub}} with scan context~\mbox{\cite{kim2018scan}}
for loop closure.

\subsubsection{Driving}
With the dual antenna RTK-GPS installed on the car roof, we obtained a precise
RTK-GPS position as the ground-truth. To ensure the validity of Global Navigation Satellite System
(GNSS), the status of the calculated position was also recorded with the $/bestpos$ topic.

%--------------------------------------------------------------------------%

\subsection{Calibration}

\label{section:calibration}

For calibration results, we offered a chain of relative poses between sensors. The sensor
chains were connected to the RGB(-D) camera for both handheld and driving sensor systems,
except for the on-board IMU sensors. More details of sensor layout, sample results, sequence statistics,
and the validity of calibration can be found on the dataset
homepage: \url{https://visibilitydataset.github.io/}.

\subsubsection{Calibration between cameras and external IMU}
The relative poses between the cameras and an external IMU were estimated by moving the sensor system
in front of the checkerboard and the grid of AprilTag~\cite{5979561}, and calibrated with
Kalibr \cite{furgale2012continuous}. Also the temporal offset
between the RGB camera and the IMU was estimated by correlating the angular velocity norms of the
sensors. The results of extrinsic calibration are shown in \figref{fig:extrinsic_result}.

\input{src/figtex/calib_results.tex}
\input{src/figtex/calib.tex}

\subsubsection{Calibration of thermal camera}
Because the thermal camera only captures the temperature but not the intensity difference,
a special type of checkerboard was required to make the pattern detectable for both
intensity and thermal cameras. We used a calibration board made of aluminum coating on the printed
circuit board (PCB) as in \figref{fig:checkerboard}. To produce detectable edges in the thermal
camera, the board was heated before calibration. Because the heat dissipation of aluminum
is higher than plastic, the aluminum pattern is cooled faster and showed a lower temperature.
Then the checkerboard pattern became visible from both RGB and thermal cameras as in \mbox{\figref{fig:thermalboard}},
enabling the thermal camera to be calibrated as the same procedure of the RGB camera.

\subsubsection{Calibration of event cameras}
The DAVIS240C event camera in the handheld sequences produces an intensity image, enabling
image-based calibration. However, in the driving sequences, the DVXplorer event camera did not
have images. Thus we chose to use reconstructed intensity images from events,
using E2VID~\mbox{\cite{Rebecq19cvpr}}. With the reconstructed AprilTag images as in \mbox{\figref{fig:eventrecon}},
the event camera could be calibrated as the same way RGB camera was calibrated.

\subsubsection{Calibration between LiDAR and camera}
The RGB camera in the handheld sequences contains depth measurements from structured light.
Thus the transformation between a LiDAR and a camera could be directly computed by running \ac{ICP}
\cite{besl1992method} on the pointclouds obtained from both sensors. The transformation is obtained by
matching planes and edges from each sensor. For best results, we have collected measurements
at the intersection of three orthogonal planes for calibration. The relative poses of sensors
were calculated by aligning orthogonal plane points.

In the driving sequences, the extrinsic parameter between a LiDAR and a RGB camera was found by
minimizing reprojection errors of the marker board plane extracted from a LiDAR and a camera.
The extrinsic transformation is $4 \times 4$ homogenous transform matrix $\mathbf{T}^{rgb}_{lidar}$,
that transforms all points from the LiDAR to the RGB frame. We captured $i$ numbers of snapshots
with fully visible checkerboard on both the camera and the LiDAR. Then, the extrinsic calibration was
calculated with both point-wise Euclidean and plane vector distance. The distances are defined
as $\mathbf{D}_{point}=\norm{\mathbf{\hat{n}}_{rgb}\cdot(\mathbf{p}_{rgb}-\mathbf{T}^{rgb}_{lidar}\mathbf{p}_{lidar})}$ and
$\mathbf{D}_{vector}=d(\mathbf{\hat{n}}_{rgb},\mathbf{T}^{rgb}_{lidar}\mathbf{\hat{n}}_{lidar})$,
where $\mathbf{p}_{rgb}$, $\mathbf{\hat{n}}_{rgb}$ are plane points and normals from the RGB camera,
and $\mathbf{p}_{lidar}$, $\mathbf{\hat{n}}_{lidar}$ are plane points and normals from the LiDAR, and
$d$ is a quaternion distance. Then the extrinsic could be found by solving the following minimization problem:
$\argmin_{\mathbf{T}^{rgb}_{lidar}} \sum_{i} (\mathbf{D}_{i, point} + \mathbf{D}_{i, vector})$.
The temporal offset between a LiDAR and sensor system was calculated by comparing the trajectory obtained
from LiDAR and visual odometry. Given the extrinsic calibration between LiDAR and camera from the
method mentioned above, the temporal offset was estimated by minimizing the Average Trajectory Error
(ATE) \mbox{\cite{Zhang18iros}} upon different temporal offset values.

\subsubsection{Calibration between marker ground-truth and sensor system}
The calibration between the motion capture system and the camera was achieved
by comparing the marker's 6DoF path of the sensor measurements. We defined a virtual
marker frame using four markers, and compared the marker frame trajectory with the
camera poses estimated by recording a checkerboard. The relative transformation between
the marker frame and the camera was obtained by following
Hand-Eye Calibration~\cite{daniilidis1999hand}. We defined the camera frame
and the marker frame at time $t$ as $\mathbf{T}_{Ct}$ and $\mathbf{T}_{Mt}$, and the
transformation of the marker frame from the camera as $\mathbf{T}_{CM}$. Then we solve the
following equation:
$\mathbf{T}^{~}_{Ci}\mathbf{T}^{~}_{CM}\mathbf{T}^{-1}_{Mi} = \mathbf{T}^{~}_{Cj}\mathbf{T}^{~}_{CM}\mathbf{T}^{-1}_{Mj}$.
Also, then using relative poses $\mathbf{T}_{Lij} = \mathbf{T}^{-1}_{Lj}\mathbf{T}^{~}_{Li}$ and
$\mathbf{T}_{Mij} = \mathbf{T}^{-1}_{Mj}\mathbf{T}^{~}_{Mi}$, the equation is reformulated into
$\mathbf{T}_{Cij}\mathbf{T}_{CM}=\mathbf{T}_{CM}\mathbf{T}_{Mij}$. Then we obtain the form of
solving $A=\mathbf{T}B\mathbf{T}^{-1}$ for $n$ pairs, and find error-minimizing
transformation $\mathbf{T}_{CM}$.
The temporal offset between ROS and the motion capture system was calculated for each sequence, by
correlating the angular velocity norms of the IMU and the motion capture poses~\mbox{\cite{furrer2018evaluation}}.

\input{src/figtex/samplealg.tex}
\newpage

%% file: src/format_table.tex
\begin{table*}[t]
	\centering
	\captionsetup{font=footnotesize}
	\caption{Sensor specifications and data types}
	\vspace{-2mm}
	\begin{tabular}{cllcc}
		\hline\hline
		Sensors & Specifications & Topic name & Description& Message type \\ \hline
		
		Thermal  & \begin{tabular}[c]{@{}l@{}}FLIR A65\\ 640\texttimes 512 pixel, 30Hz / 20Hz (handheld)\\ FOV : 90\degree  vert., 69\degree  horiz.\\ Spectral Range : 7.5-13 $\mu$m\end{tabular} & \begin{tabular}[c]{@{}l@{}}/thermal/camera\_info\\ /thermal/image\_raw\end{tabular}                                                                     & \begin{tabular}[c]{@{}c@{}}Header\\ Image (16bit, 1ch)\end{tabular}                             & \begin{tabular}[c]{@{}c@{}}sensor\_msgs/CameraInfo\\ sensor\_msgs/Image\end{tabular}                                                 \\ \hline
		
		RGB-D    & \begin{tabular}[c]{@{}l@{}}Asus Xtion Pro Live\\ 640\texttimes 480 pixel, 30Hz\\ FOV : 45\degree  vert., 58\degree  horiz.\end{tabular} & \begin{tabular}[c]{@{}l@{}}/depth/camera\_info\\ /depth/image\_raw\\ /rgb/camera\_info\\ /rgb/image\end{tabular} & \begin{tabular}[c]{@{}c@{}}Header\\ Image (16bit, 1ch)\\ Header\\ Image (8bit, 3ch)\end{tabular} & \begin{tabular}[c]{@{}c@{}}sensor\_msgs/CameraInfo\\ sensor\_msgs/Image\\ sensor\_msgs/CameraInfo\\ sensor\_msgs/Image\end{tabular} \\ \hline
		
		RGB    & \begin{tabular}[c]{@{}l@{}}Flea®3 FL3-U3-13E4C-C \\ 1280\texttimes 1024 pixel, 60Hz\\FOV : 89\degree  vert., 73.8\degree  horiz.\end{tabular} & \begin{tabular}[c]{@{}l@{}}/camera/camera\_info\\/camera/image\_color\end{tabular} & \begin{tabular}[c]{@{}c@{}}Header\\ Image (8bit, 3ch)\end{tabular} &\begin{tabular}[c]{@{}c@{}}sensor\_msgs/CameraInfo\\ sensor\_msgs/Image\end{tabular} \\ \hline
		
		Event    & \begin{tabular}[c]{@{}l@{}}Handheld : DAVIS 240C\\ 240\texttimes 180 pixel, upto 12 MEPS\\Driving : DVXplorer\\ 640\texttimes 480 pixel, upto 165 MEPS\\ \end{tabular} & /dvs/events & Event &dvs\_msgs/EventArray \\ \hline
	
		GPS & \begin{tabular}[c]{@{}l@{}}NovAtel Pwrpak7 E-1 \\ VEXXIS GNSS-500 Dual Antenna\end{tabular}
		& \begin{tabular}[c]{@{}l@{}}/gps \\ /bestpos\end{tabular}
		& GPS
		& \begin{tabular}[c]{@{}c@{}}gps\_common/GPSFix \\ novatel\_gps\_msgs/NovatelPosition\end{tabular} \\ \hline
	
		Inertial &\begin{tabular}[l]{@{}l@{}}LORD Microstrain 3DM-GX5-25\\On-board IMU (MPU 6150)\\ On-board IMU (MPU 9250)\\ On-board IMU (InvenSense ICM-20948)\end{tabular}
		& \begin{tabular}[c]{@{}l@{}}/imu/data \\/dvs/imu \\/dvs/imu\\ /os1\_cloud\_node/imu\end{tabular} & IMU & sensor\_msgs/Imu  \\ \hline
		
		LiDAR &\begin{tabular}[l]{@{}l@{}}Handheld : Velodyne VLP-16\\ Driving : Ouster OS1-64\end{tabular}
		& \begin{tabular}[c]{@{}l@{}}/velodyne\_point\_cloud\\ 
		/os1\_cloud\_node/points\end{tabular}                                                                                            &     Pointcloud                               & sensor\_msgs/PointCloud2 \\ \hline
	\end{tabular}
\vspace{-5mm}
\label{tab:types}
\end{table*}

%% file: src/figtex/calib_results.tex
\setcounter{figure}{1}

\begin{figure}[h]
	\centering
	\captionsetup{font=footnotesize}
	\subfigure[RGB-D pointcloud projected \newline to event frames in the handheld seq.]{%
		\includegraphics[width=0.5\columnwidth]{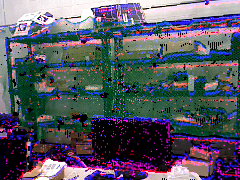}
		\label{fig:handheld_ext}
		\hspace{-3mm}
	}%
	\subfigure[LiDAR pointcloud projected \newline to RGB frames in the driving seq.]{%
		\includegraphics[width=0.5\columnwidth,trim = 0 20 0 20, clip]{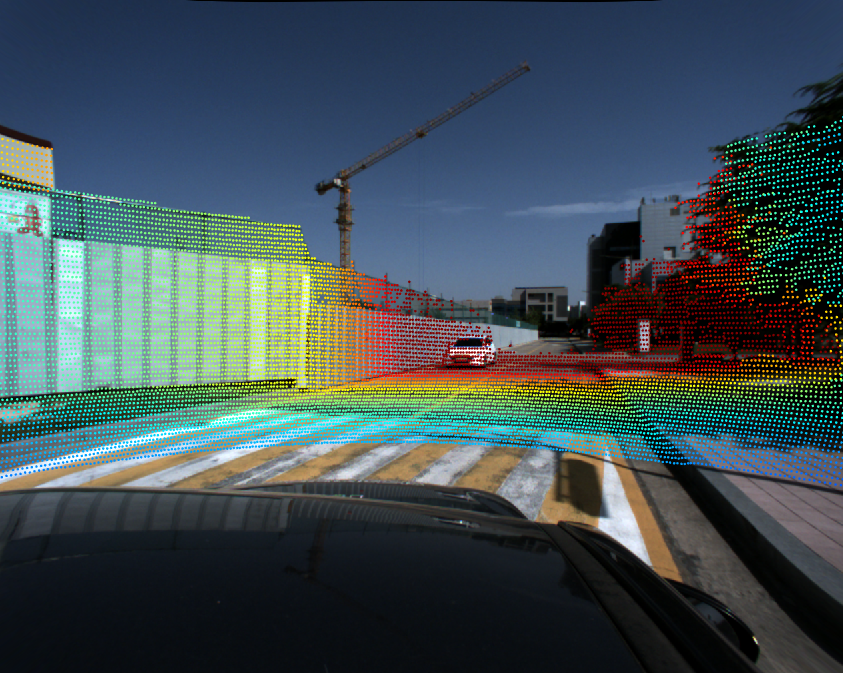}
		\label{fig:driving_ext}
	}
	\vspace{-2mm}
	\caption{Pointclouds transformed into the other cameras, by the calibration results.
		In (b), LiDAR pointclouds are psudocolored with the depth values.}
	\label{fig:extrinsic_result}
	\vspace{-4mm}
\end{figure}

%% file: src/figtex/calib.tex
\setcounter{figure}{2}
\begin{figure}[h!]
	\centering
	\captionsetup{font=footnotesize}
	\subfigure[RGB image of \newline PCB checkerboard.]{%
		\includegraphics[width=0.33\columnwidth]{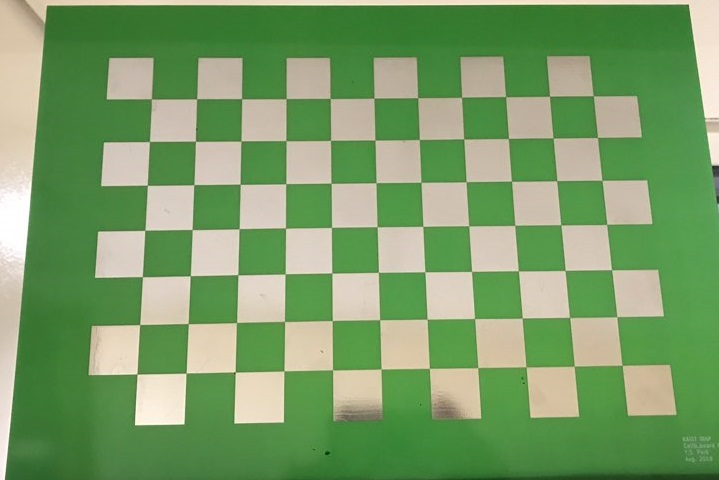}
		\hspace{-3mm}
		\label{fig:checkerboard}
	}%
	\subfigure[Thermal image of \newline heated PCB.]{%
		\includegraphics[width=0.33\columnwidth,trim = 0 40 0 42, clip]{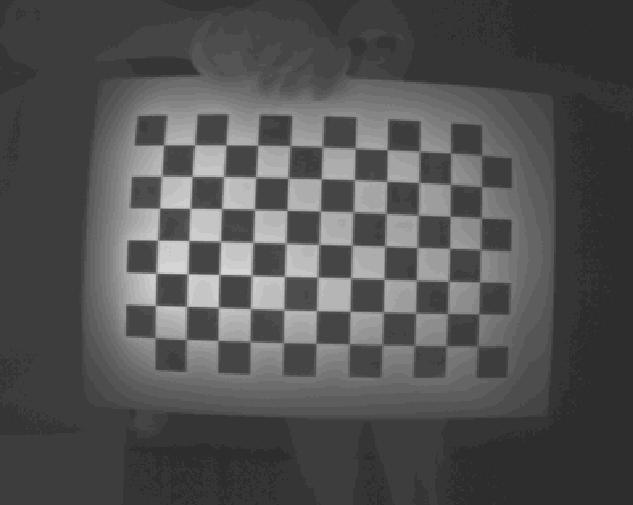}
		\label{fig:thermalboard}
		\hspace{-3mm}
	}%
	\subfigure[Image of AprilTag \newline recovered from events.]{%
		\includegraphics[width=0.33\columnwidth,trim = 0 28 0 25, clip]{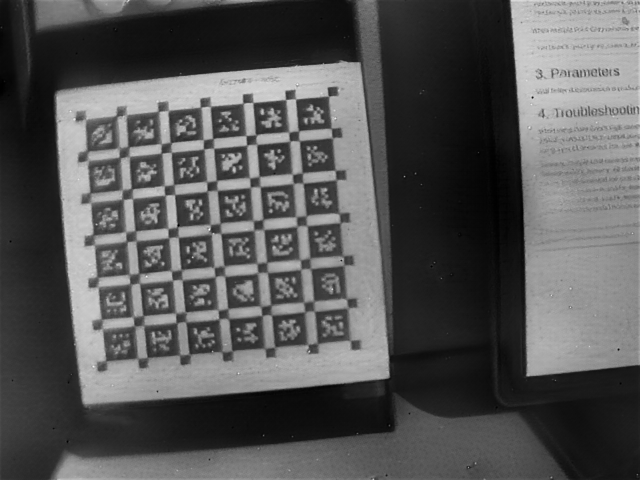}
		\label{fig:eventrecon}
	}

	\vspace{-2mm}
	\caption{Checkerboard pattern used for extrinsic calibration between alternative vision sensors.
		In (b), the difference in the heat dissipation rate of each material results in a temperature pattern.
		In (c), an intensity image is reconstructed from events using E2VID~\mbox{\cite{Rebecq19cvpr}} and used for calibration.
	}
\vspace{-5mm}
\end{figure}

%% file: src/figtex/samplealg.tex
\begin{figure}[!h]
	\centering
	\vspace{-1mm}
	\captionsetup{font=footnotesize}
	\subfigure[PR curves of campus-day1 seq. \newline matched to campus-day2 DB.]{%
		\captionsetup{justification=centering}
		\includegraphics[width=0.5\columnwidth]{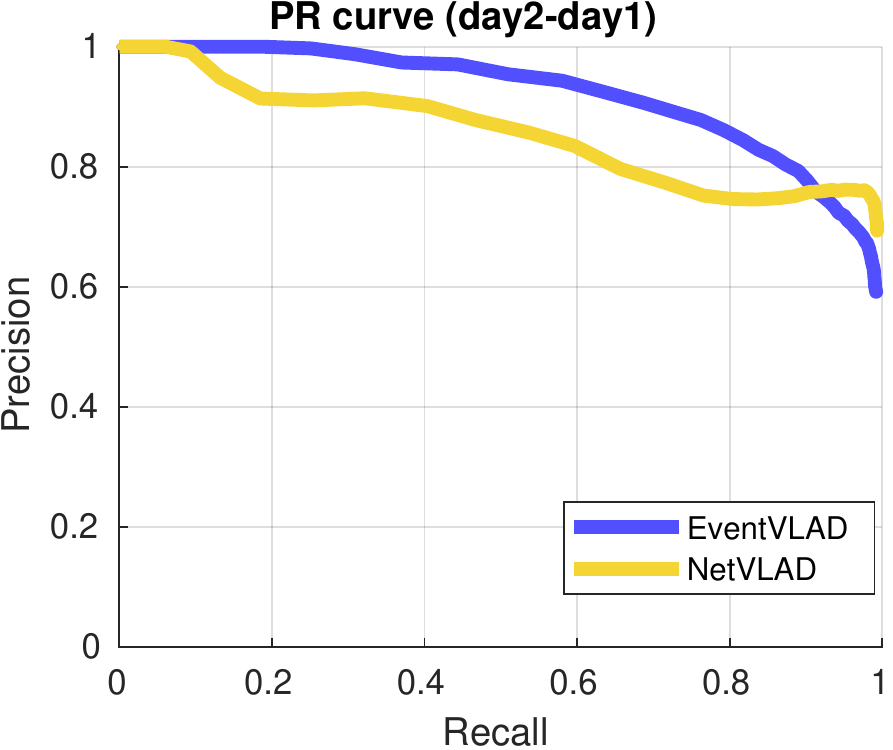}
	}%
	\subfigure[PR curves of campus-night seq. \newline matched to campus-day2 DB.]{%
		\includegraphics[width=0.5\columnwidth]{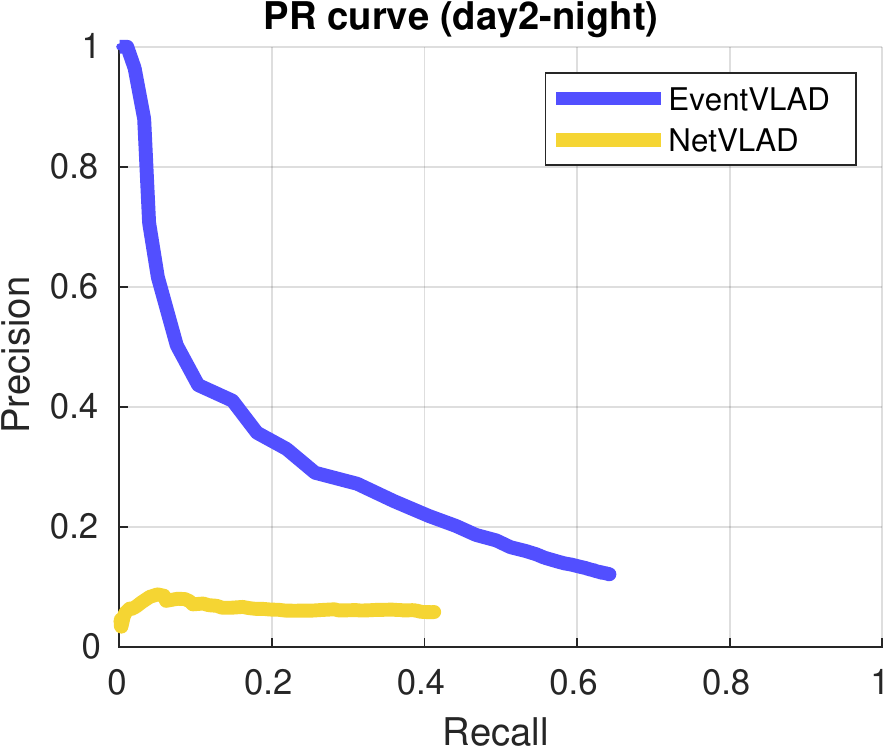}
	}
	\vspace{-1mm}
	\caption{VPR performances of event(blue) and RGB(yellow) cameras, trained with images
		from all of other ``day" sequences and tested on the Campus sequences. By consisting
		a DB only with Campus-Day2, we tested the robustness of VPR module to match
		Campus-Day1/Night images to DB. The event-based
		VPR has shown more robustness over appearance changes.
	}
	\label{fig:extrinsic_result_drive}
	\vspace{-1mm}
\end{figure}

%% file: src/dataset.tex
\section{Dataset}

We recorded multiple repetitions along similar trajectories under distinct illumination
conditions and motion constraints. To show the robustness of alternative vision sensors
over illumination conditions, we have shown the performance of event-based VPR~\cite{lee2021event}
in \figref{fig:extrinsic_result_drive}. To overview the sequences, 3D reconstructions and samples are in the figures:
Figs. \hyperref[fig:seq_handheld]{5} and \hyperref[fig:samples_handheld]{6}
for handheld and Figs. \hyperref[fig:seq_driving]{7} and \hyperref[fig:samples_driving]{8} for driving.

\subsection{Handheld sequences}

\input{src/seq_list_handheld.tex}

For handheld sequences, we recorded multiple sessions in indoor and outdoor
environments, respectively. In \figref{fig:seq_handheld}, the trajectory and
sample scene images are displayed. The full sequence list is detailed in \tabref{tab:sequences_1}.

For handheld indoor, sequences were divided by the level of motion deviation and lighting
conditions. We have repeated trajectories at three different motion levels (42.7 to 120.7\mbox{$^{\circ}$}/s), which are slow,
unstable, and aggressive. Further, for each motion, we changed the illumination condition, from
light turned on (global), turned off (dark), and flashlight on (local). We also included a
bonus sequence (varying) to gradually change the flashlight intensity from dark to local. The sensor system's global poses were
captured via a motion capture system with 12 motion capture cameras mounted on the wall. The scale
of the room was \unit{12.3}{m} $\times$ \unit{8.9}{m}.

For handheld outdoor, we trailed the same trajectory at different times clockwise and
counterclockwise. The approximate size of the space was about \unit{60}{m} $\times$ \unit{40}{m}.
In the environment, GPS was not sufficient because multi-story buildings enclosed the experiment
site and yielded a multipath reception problem~\mbox{\cite{kos2010effects}}. However, we were able to obtain
LiDAR odometry from the plenty of planar and edge structures at the building.

\subsection{Driving sequences}

\input{src/seq_list_driving.tex}

For driving sequences, we repeated measurements along the predefined trajectories at different
times. The repetitions were done throughout the entire day and night to enclose sun elevation changes.
The appearances of the same place broadly change at different times, as seen in \figref{fig:seq_driving}.
The path was carefully selected to include visual loop closure, and the recording was done
repeatedly over identical paths.

We set up three different trajectories, around the campus, in the urban canyon, and along the city's
arterial road. Each of the sequences was named as Campus, City, and Urban in \tabref{tab:sequences_2}.
The campus sequence consisted of pedestrians, bicycles, and medium-rise buildings. In the city
sequences, the vehicle drives along the river, under and over the bridges, and underpasses. In
the urban sequences, the sensor observed high-rise buildings and tall trees (urban canyon). To
provide extra data for unsupervised or self-supervised learning, we prepared ``Vision" sequences recorded only with cameras
without a LiDAR. To ensure reproducibility, we recommend to use the Campus sequences for validation,
as in the setup of \figref{fig:extrinsic_result_drive}. Each traverse was from 3.6 to 9.3km long,
with a maximum linear and angular velocity of 95~km/h and 39~$^{\circ}$/s. As the sun elevation angle changes, shadows differ, and large
color shifts occurred in the evening scenes. Further, at night, artificial lights, including street lamps,
were turned on. These illumination changes made the urban city very different from itself along the time of
the day. Although the vehicle followed the same trajectory, the average speed dropped to 65\% by the traffic
signals and traffic congestion.

\section{Issues}

The spectral response of the DAVIS240C used for our experiment overlaps with the infrared
wavelength used from Cortex motion capture system. Thus, directly using DAVIS240C in the
motion capture room resulted in an event camera's high malfunction, capturing all the flashes
of infrared strobes and ruining the event measurement. Therefore, we attached a C-Mount IR cut
filter in front of the event camera to remove the effects of motion capture strobes.

The event cameras possess very high sensitivity and provide a high dynamic range from their relative
form of luminance illustration. However, due to the sensitivity, we identified that the event camera
produced erroneous output and decreased data quality when directly observing the light source.
As is shown in the last column of \figref{fig:samples_driving}, the event camera shows
artifacts from the street lamps and loses visual information near the lamp. Further studies are
required to neglect artifacts and recover the dynamic range of events.

\input{src/figtex/seq_handheld.tex}

\input{src/figtex/seq_driving.tex}
\newpage

%% file: src/seq_list_handheld.tex
% \usepackage{multirow}
\begin{table}[b]
	\centering
	\vspace{-3mm}
	\captionsetup{font=footnotesize}
	\caption{Environment setting for handheld sequences}
	\vspace{-1mm}
	\begin{tabular}{ccccc}
		\hline
		Location                 & Illumination & Motion & Duration &  GT \\ \hline
		\multirow{10}{*}{Indoor} 
		& \multirow{3}{*}{Global}  & Slow           & 52.5s &  \multirow{10}{*}{Vicon} \\ \cline{3-4} 
		&                          & Unstable       & 23.9s &  \\ \cline{3-4} 
		&                          & Aggressive     & 16.9s &  \\ \cline{2-4} 
		& \multirow{3}{*}{Dark}    & Slow           & 35.0s &  \\ \cline{3-4} 
		&                          & Unstable       & 20.5s &  \\ \cline{3-4} 
		&                          & Aggressive     & 16.9s &  \\ \cline{2-4} 
		& \multirow{3}{*}{Local}   & Slow           & 35.0s &  \\ \cline{3-4} 
		&                          & Unstable       & 17.1s &  \\ \cline{3-4} 
		&                          & Aggressive     & 13.4s &  \\ \cline{2-4} 
		& Varying                  & Slow           & 35.8s &  \\ \hline
		\multirow{4}{*}{Outdoor}
		& Day1               & Slow         & 117s   & \multirow{4}{*}{\begin{tabular}{cc}LOAM\\(baseline)\end{tabular}} \\ \cline{2-4} 
		& Day2               & Slow         & 108s   &\\ \cline{2-4} 
		& Night1             & Slow         & 101s   &\\ \cline{2-4} 
		& Night2             & Slow         & 104s   &\\ \hline
	\end{tabular}
\label{tab:sequences_1}
\vspace{-1mm}
\end{table}

%% file: src/seq_list_driving.tex
\begin{table}[b]
	\centering
	\captionsetup{font=footnotesize}
	\vspace{-3mm}
	\caption{Environment setting for driving sequences}
	\vspace{-1mm}
	\begin{tabular}{ccccc}
		\hline
		Type & Location            & Time        & Duration  & GT \\ \hline
		\multirow{8}{*}{\begin{tabular}{cc}Vision\\+LiDAR\end{tabular}} & \multirow{4}{*}{\begin{tabular}{cc}Campus\\(3.6km)\end{tabular}}
		& Day1        &    445s  &  \multirow{8}{*}{\begin{tabular}{cc}GPS, \\LOAM\end{tabular}} \\ \cline{3-4}
		&               & Day2        &    463s  &  \\ \cline{3-4}
		&               & Evening     &    438s  &  \\ \cline{3-4}
		&               & Night       &    445s  &  \\ \cline{2-4}
		& \multirow{4}{*}{\begin{tabular}{cc}City\\(9.3km)\end{tabular}}
		& Day1        &    914s &  \\ \cline{3-4}
		&               & Day2        &    1218s&  \\ \cline{3-4}
		&               & Evening     &    832s &  \\ \cline{3-4}
		&               & Night       &    840s &  \\ \hline
		\multirow{14}{*}{Vision} & \multirow{5}{*}{\begin{tabular}{cc}Campus\\(3.6km)\end{tabular}}
		& Morning     &    495s &  \multirow{14}{*}{GPS} \\ \cline{3-4}
		&               & Day1        &    667s &  \\ \cline{3-4}
		&               & Day2        &    484s &  \\ \cline{3-4}
		&               & Evening     &    487s &  \\ \cline{3-4}
		&               & Night       &    443s &  \\ \cline{2-4}
		& \multirow{5}{*}{\begin{tabular}{cc}City\\(9.3km)\end{tabular}}
		& Morning     &    858s &  \\ \cline{3-4}
		&                 & Day1        &    1037s&  \\ \cline{3-4}
		&                 & Day2        &    890s &  \\ \cline{3-4}
		&                 & Evening     &    1127s&  \\ \cline{3-4}
		&                 & Night       &    855s &  \\ \cline{2-4}
		& \multirow{4}{*}{\begin{tabular}{cc}Urban\\(3.8km)\end{tabular}}
		& Morning     &    652s &  \\ \cline{3-4}
		&               & Day         &    785s &  \\ \cline{3-4}
		&               & Evening     &    593s &  \\ \cline{3-4}
		&               & Night       &    598s &  \\ \hline
	\end{tabular}
	\label{tab:sequences_2}
	\vspace{-1mm}
\end{table}

%% file: src/figtex/seq_handheld.tex
\begin{figure*}[!h]
    \centering
    \vspace{2mm}
    \includegraphics[width=\textwidth]{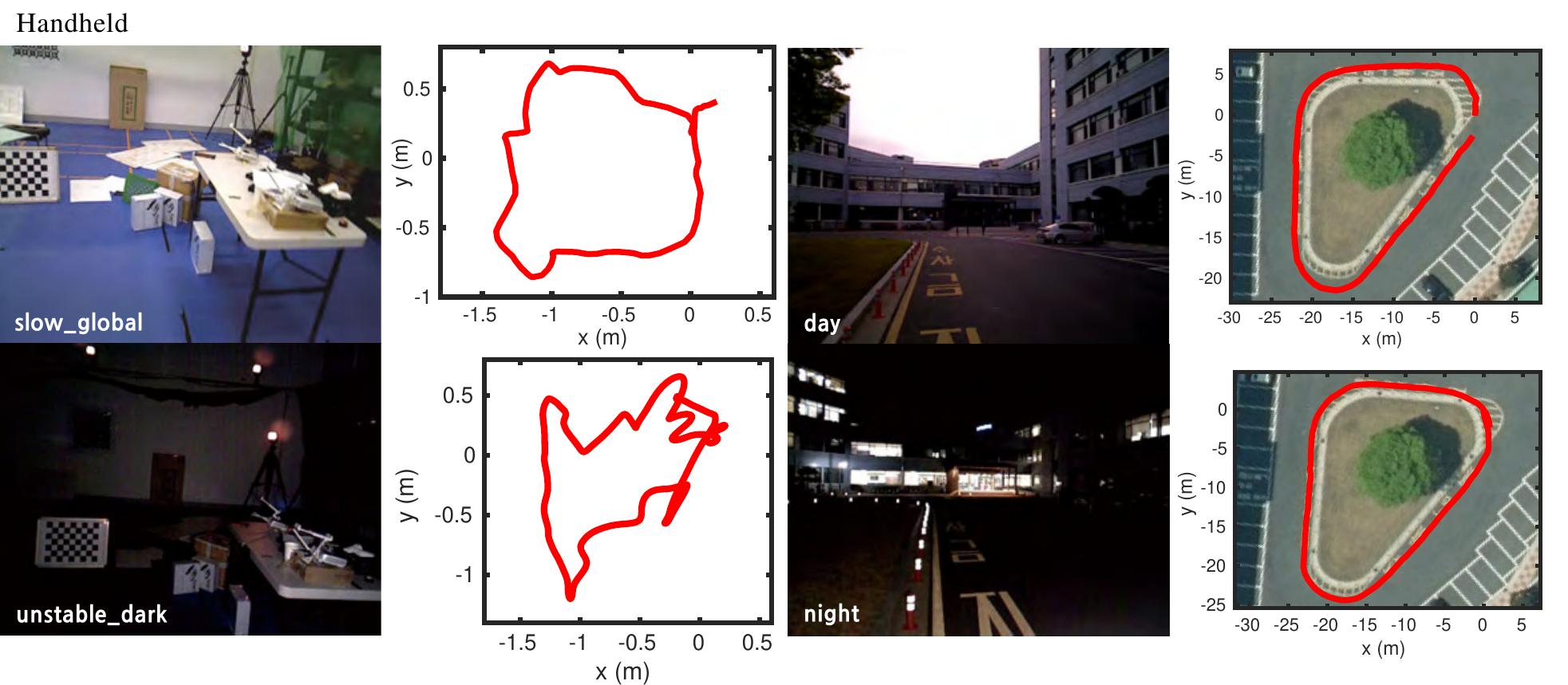}

	\captionsetup{font=footnotesize}
    \caption{The samples of appearance changes and corresponding ground-truth or baseline
    	trajectories (motion capture for indoor, SLAM for outdoor) from handheld
    	sequences. For indoor trajectories, the measurements were done in the
    	same room, but with different illumination and motion speed. Outdoor
    	 sequences were recorded by walking around the
    	 environment in a clockwise and counter-clockwise
    	 direction at different times.}

    \label{fig:seq_handheld}
    
	\centering
	\vspace{3mm}
	\includegraphics[width=\textwidth]{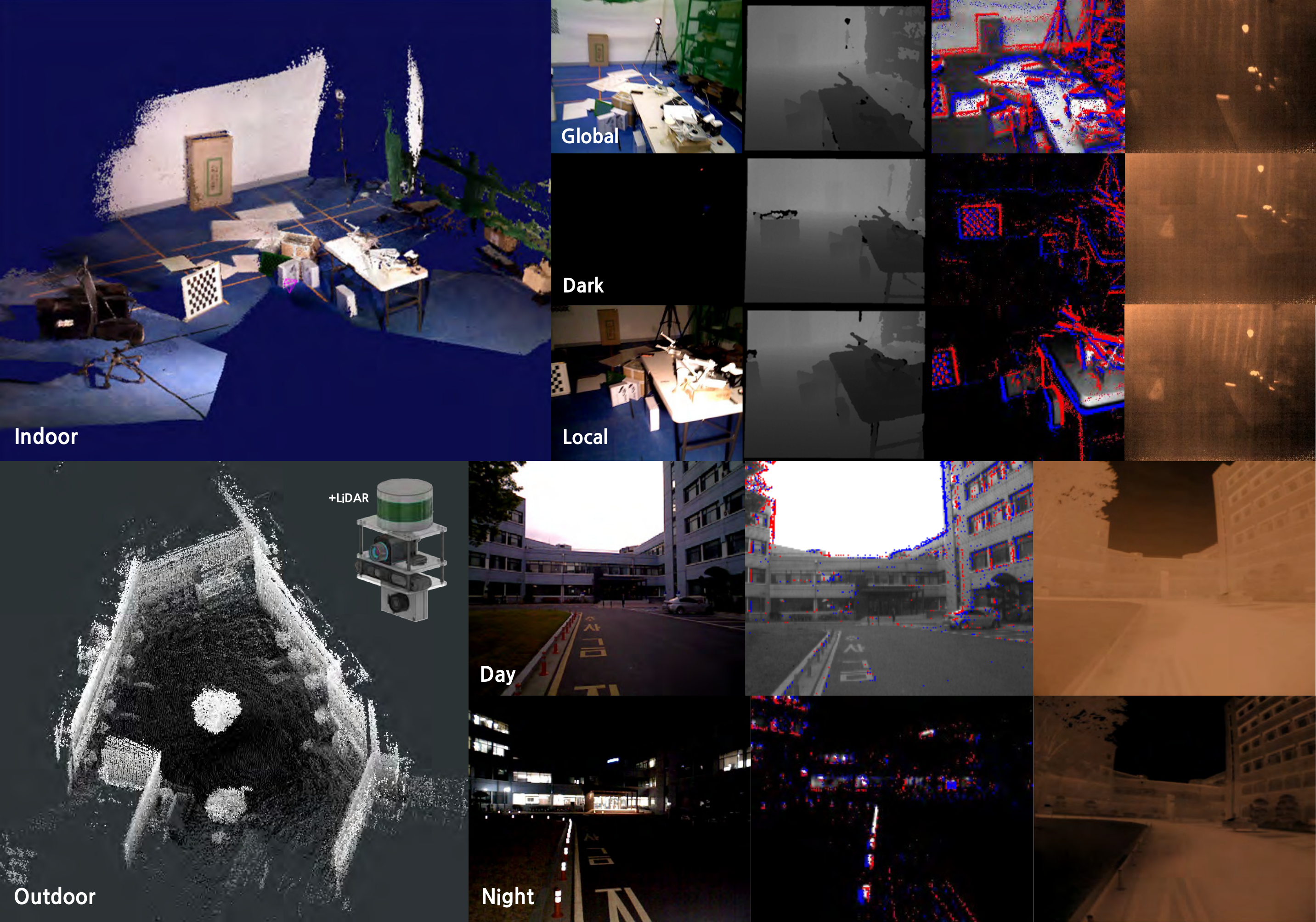}
	
	\caption{Handheld scene overview with 3D reconstruction of the environment.
		From left to right, sampled images from RGB, depth, event and thermal camera are shown.
		The text label at each row represents illumination condition. The indoor environment was reconstructed
		with known poses from a motion capture and a RGB-D camera, and the outdoor was reconstructed by LeGO-LOAM
		run from VLP-16 LiDAR attached at the top of the sensor system.}
	\label{fig:samples_handheld}
	\vspace{2mm}
\end{figure*}

%% file: src/figtex/seq_driving.tex
\begin{figure*}[!t]
	\centering
	\vspace{2mm}
	\includegraphics[width=\textwidth]{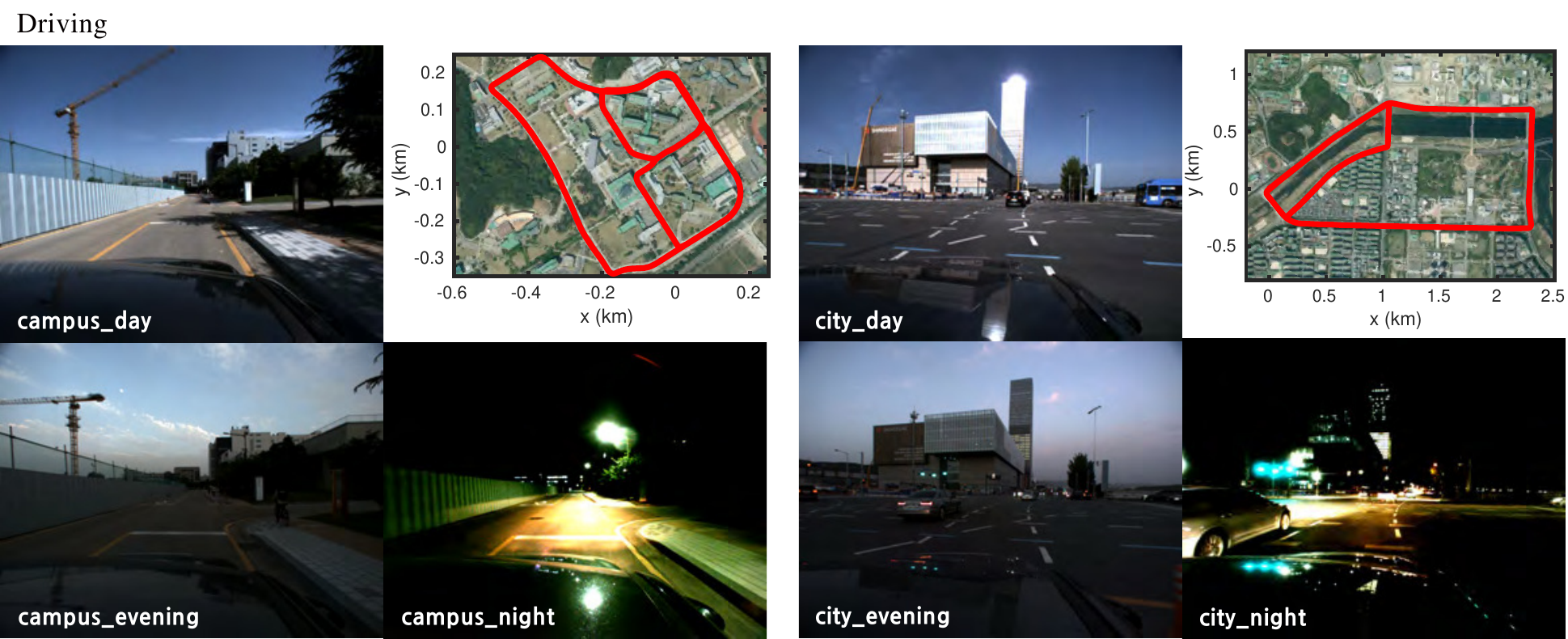}
	\captionsetup{font=footnotesize}
	\caption{The samples of appearance changes and corresponding GPS trajectories from Vision for
		Visibility Dataset driving sequences. Outdoor sequences were recorded
		by following predefined path, at different times. Due to the illumination changes, visible appearance
		largely shifts as in the figure.}
	\label{fig:seq_driving}
	
	\centering
	\vspace{2mm}
	\includegraphics[width=\textwidth]{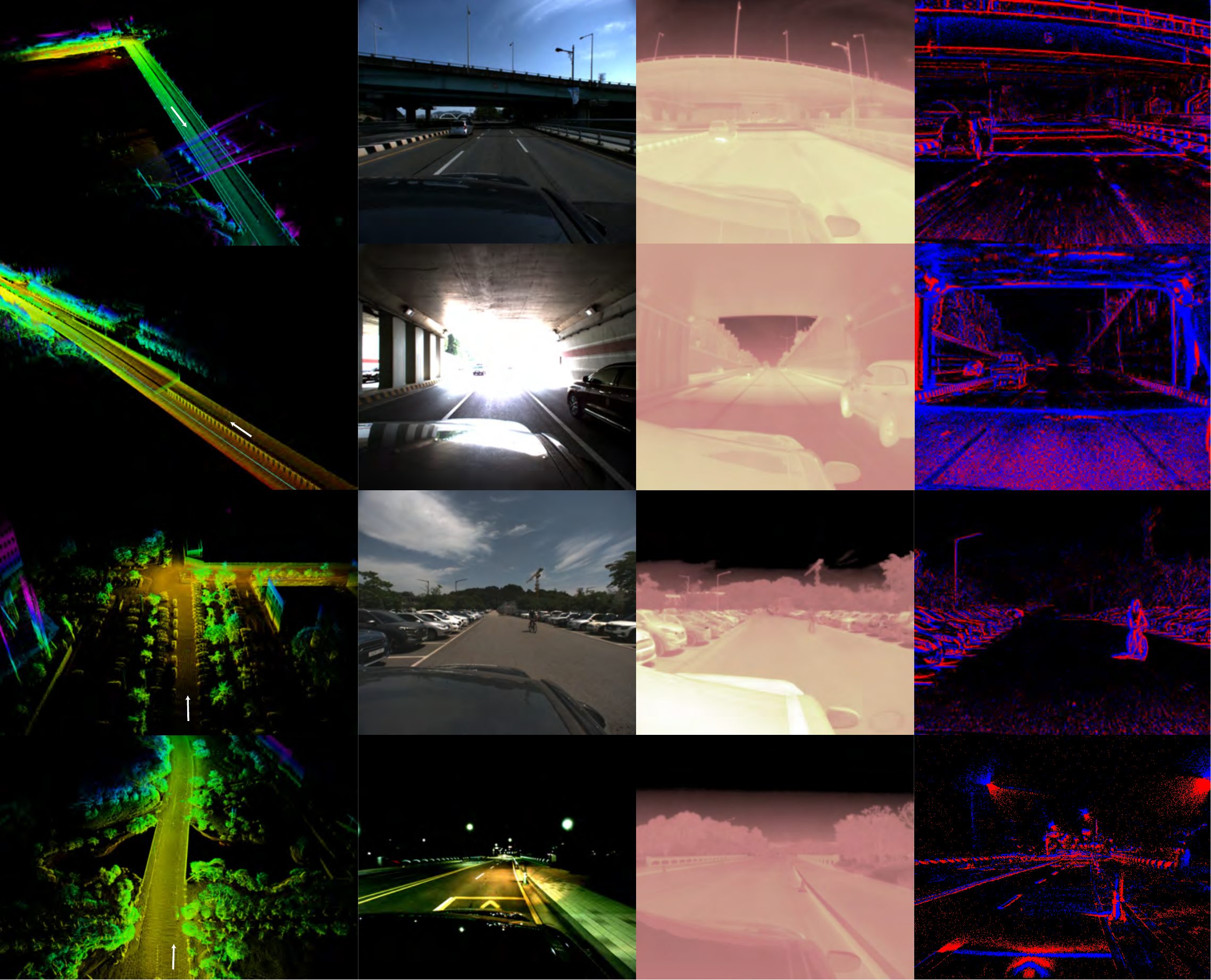}
	\caption{Vision+LiDAR sequences of City and Campus, with reconstruction from A-LOAM and the current viewpoint indicated with a white arrow. In the city, road infrastructures such as overpass and underpass / high-rise and large buildings / riverside are included. In the campus, parked cars / vegetation / medium-sized buildings / construction sites are observed. We could observe the light artifacts from the street lamps, in the last row of the event outputs.}
	\label{fig:samples_driving}
\end{figure*}

%% file: src/conclusion.tex
\section{Conclusion}

This paper provides a vision for visibility dataset to overcome poor lighting conditions in robotics
applications: beyond the visible light spectrum and temporal luminance differences. Besides
providing a reference environment to test the performance of the thermal and event cameras, our dataset
enables researchers to determine the vision sensors' required abilities for robotic applications:
fully identifying the potential of passive cameras in the real world. We hope our work will help
solve robust robot vision regardless of motion or environmental disturbances by encouraging
visual SLAM based on alternative vision sensors, proposing a test-bed to test optimal camera
characteristics in the real world.